\documentclass[conference,12pt,onecolumn,draftclsnofoot]{IEEEtran}

\usepackage[utf8]{inputenc}
\usepackage{geometry}
\usepackage{graphicx}
\usepackage{booktabs}
\usepackage{hyperref}
\usepackage{pgfplots}
\usepackage{float}
\usepackage{cite}

\hypersetup{
    colorlinks=true,
    linkcolor=blue,
    filecolor=magenta,      
    urlcolor=cyan,
    citecolor=blue,
}

\usepackage{xcolor}
\usepackage{pgfplots}
\pgfplotsset{compat=1.18}

\definecolor{cb-orange}{HTML}{E69F00}
\definecolor{cb-sky}{HTML}{56B4E9}
\definecolor{cb-green}{HTML}{009E73}

\usepackage{tabularx}
\usepackage{booktabs}

\usepackage{microtype}
\usepackage{color}
\usepackage{xspace}
\usepackage{amsmath}
\usepackage{amssymb}
\usepackage{amsthm}
\usepackage{tikz}
\usepackage{url}
\usepackage{enumitem}

\usepackage{titlesec}
\renewcommand{\thesection}{\arabic{section}}
\renewcommand{\thesubsection}{\arabic{section}.\arabic{subsection}}
\renewcommand{\thesubsubsection}{\arabic{section}.\arabic{subsection}.\arabic{subsubsection}}

\titleformat{\section}
  [hang]
  {\normalfont\bfseries\Large}
  {\thesection.}
  {1em}
  {}

\titleformat{\subsection}
  [hang]
  {\normalfont\bfseries\large}
  {\thesubsection.}
  {1em}
  {}

\titleformat{\subsubsection}
  [hang]
  {\normalfont\bfseries}
  {\thesubsubsection.}
  {1em}
  {}

\title{A Persona-Based Evaluation Framework for Pluralistic Alignment in Generative AI}

\author{
    \IEEEauthorblockN{\large Atahan Karagöz}
    \IEEEauthorblockA{
        \textit{Department of Computer Science} \\
        \textit{University of Basel} \\
        Basel, Switzerland \\
        atahan.karagoez@stud.unibas.ch
    }
}

\begin{document}

\maketitle

\begin{abstract}
\noindent Current alignment paradigms for generative artificial intelligence rely predominantly on monolithic benchmarking frameworks that reduce the plurality of human judgment to aggregated statistical baselines, thereby obscuring cultural, demographic, and contextual variability in evaluation. We introduce a state-space constrained emulation framework for AI evaluation that replaces singular assessment functions with a structured manifold of synthetic cognitive profiles representing diverse human perspectives. We show that modern generative architectures can instantiate and maintain these evaluative personas with high consistency, enabling a form of pluralistic, perspective-dependent benchmarking that more closely reflects real-world consensus variability. However, we further analyze the stability of these simulated evaluators under sequential inference and stochastic prompt perturbations, revealing systematic degradation in persona coherence that manifests as state-space drift and semantic inconsistency. These findings suggest that static alignment constraints are insufficient for sustaining robust evaluative behavior over time. Instead, we argue for the necessity of embedding dynamic, viability-driven regulatory mechanisms within generative systems to preserve coherent cognitive emulation. By framing persona-based evaluation as a structured dynamical system over latent representation manifolds, this study provides a foundation for more adaptive, human-aligned, and context-sensitive approaches to AI evaluation.
\end{abstract}


\section{Introduction}
\label{sec:intro}

The emergence of highly capable generative artificial intelligence has fundamentally altered the paradigm of digital synthesis. Modern multimodal architectures have evolved from narrow functional operators into expansive systems capable of simulating complex, intersectional realities \cite{vaswani2017attention, lecun2015deep}. However, as these models achieve unprecedented technical proficiency, the methodologies utilized to evaluate them have reached a critical bottleneck \cite{brown2020language, kaplan2020scaling, ouyang2022training}. Traditional benchmarking predominantly relies on monolithic automated metrics or a singular, homogenized ``AI-as-a-judge'' framework. These objective approaches inherently fail to capture the sociological nuances, pluralistic values, and cultural subjectivity that define authentic human judgment.

The integration of generative models into global workflows reveals a profound alignment challenge. While they exhibit remarkable creative utility, true AI alignment cannot be treated as a single, universal vector. An AI system evaluated through a homogenized algorithmic lens often collapses into culturally stereotyped or averaged representations, frequently amplifying the latent biases embedded within its vast training corpora \cite{bender2021dangers, buolamwini2018gender}. In our foundational analysis of generative content creation, we established that models exhibit high algorithmic adherence but lack the rigorous contextual constraints necessary for diverse human application \cite{karagoz2024ethicstechnicalaspectsgenerative}. Consequently, robust AI evaluation requires a framework as diverse and intersectional as the populations it serves.

To transcend the limitations of a monolithic viewpoint, recent literature has pivoted toward utilizing language models as interactive simulacra \cite{park2023generative, shanahan2023role}. Building upon this, this paper introduces a comprehensive persona-based evaluation paradigm. We investigate the capacity of generative systems to emulate distinct demographic, professional, and cultural identities, utilizing these simulated perspectives to subjectively score artificial outputs. Crucially, the capability of an artificial agent to sustain a coherent, highly specific persona over extended evaluative iterations is a nontrivial challenge of state-space stability. Sustaining this localized state-space over time requires internal regulation, sharing structural parallels with principles of computational inertia \cite{karagoz2025computationalinertiaconservedquantity}. Furthermore, mapping these diverse cognitive simulations shares deep geometric parallels with the unsupervised discovery of complex, self-organizing biological manifolds and probabilistic latent spaces \cite{kingma2013auto, karagoz2025selforganizingsurvivalmanifoldstheory}. 

By shifting the evaluative lens from an objective machine baseline to a continuous manifold of simulated human subjectivities, this research bridges the gap between technical benchmarking and humanistic nuance. To comprehensively investigate this framework, this study outlines three primary objectives:
\begin{enumerate}
    \item \textbf{Quantify Persona Fidelity in Evaluation:} To assess the architectural capacity of generative models to maintain distinct, context-aware evaluative lenses across both textual and visual synthesis tasks.
    \item \textbf{Map Algorithmic vs. Pluralistic Divergence:} To systematically compare persona-driven subjective evaluations with established baseline human consensus, identifying critical points of alignment and algorithmic divergence.
    \item \textbf{Establish Theoretical Grounding for Cognitive Emulation:} To analyze patterns in persona stability, linking the practical emulation of human perspectives to broader theories of continuous-time optimization, autonomous regulation, and geometric state-space preservation.
\end{enumerate}

\section{Literature Review}
\label{sec:lit_review}

The rapid evolution of generative artificial intelligence has shifted the focus of machine learning research from syntactic synthesis to semantic and cultural alignment. The foundations of this transformation rely heavily on the Transformer architecture \cite{vaswani2017attention}, foundational deep learning principles \cite{lecun2015deep}, and the scaling laws governing compute-optimal few-shot learners \cite{brown2020language, kaplan2020scaling, hoffmann2022training, openai2023gpt4}. Parallel breakthroughs in latent diffusion models have similarly revolutionized visual synthesis, enabling systems to generate high-fidelity imagery by elucidating and optimizing the generative design space \cite{rombach2022high, karras2022elucidating}. While these models demonstrate extraordinary representational capacity, their unprecedented scale inevitably encodes systemic and societal biases \cite{bender2021dangers}. Our prior foundational investigation into generative workflows established that while models exhibit high creative utility, their outputs frequently succumb to cultural homogenization, perpetuating implicit demographic stereotypes when evaluated through a standard algorithmic lens \cite{karagoz2024ethicstechnicalaspectsgenerative}. This vulnerability underscores the necessity of transitioning from rigid, objective benchmarking to frameworks capable of capturing pluralistic subjectivities \cite{buolamwini2018gender, ouyang2022training}.

To address the limitations of homogenized evaluation, recent literature has pivoted toward utilizing large language models as interactive simulacra, capable of sustaining dynamic role-play and human-like agentic behavior \cite{park2023generative, shanahan2023role}. However, maintaining a highly specific, synthetic persona over an extended sequence of stochastic interactions introduces a profound theoretical challenge concerning state-space stability. Sustaining a coherent cognitive simulacrum is not merely a prompting task; it is fundamentally a problem of continuous-time optimization and autonomous regulation.

We can analyze the stability of persona emulation through the geometric properties of learning dynamics. During inference, the continuous constraints of an assigned persona function analogously to a conserved quantity within the model's trajectory. This preservation reflects the principles of computational inertia \cite{karagoz2025computationalinertiaconservedquantity}, where the simulated identity resists the stochastic perturbations introduced by varying user prompts, much like a frictionless optimization path resisting chaotic divergence. Sustaining this localized state-space over time requires internal viability-constrained regulation, sharing structural parallels with the thermodynamic persistence observed in energentic intelligence paradigms \cite{karagoz2025energenticintelligenceselfsustainingsystems}. In this context, the generative agent must self-regulate its internal state to prioritize the ``survival'' of the assigned persona's coherence over immediate, unconstrained token maximization. This paradigm extends traditional frameworks of self-supervised representation learning by treating continuous alignment as a dynamic objective \cite{chen2020simple}.

Furthermore, representing a diverse pluralism demands the mapping of highly heterogeneous cultural and demographic attributes into a unified, stable latent space. The mechanics of embedding such disparate modalities into distinct, cohesive clusters draw heavily upon foundational theories of variational and contrastive embeddings \cite{kingma2013auto, karagoz2025omicsclunsupervisedcontrastivelearning}. Just as unsupervised contrastive objectives, such as OmicsCL, can successfully extract and stratify heterogeneous biological modalities into distinct, clinically meaningful subtypes \cite{karagoz2025omicsclunsupervisedcontrastivelearning}, generative models must autonomously stratify vast textual distributions into discrete, self-consistent human profiles. 

Ultimately, the emergence and preservation of these simulated human perspectives within the neural architecture can be conceptualized as distinct geometric flows. The stability of these simulated evaluators mirrors the self-organizing survival manifolds found in complex biological systems, where sustainable, aligned representations naturally emerge as low-curvature geodesic paths guided by internal structural constraints \cite{karagoz2025selforganizingsurvivalmanifoldstheory}. By grounding persona-based evaluation in these theoretical physics and geometric principles, we establish a robust foundation for understanding how artificial agents can reliably emulate diverse human judgments, ultimately advancing the frontier of scalable, human-aligned AI \cite{bai2022constitutional}.

\section{Methodology}
\label{sec:methodology}

To systematically evaluate the capacity of generative models to simulate and sustain pluralistic human perspectives, we engineered a state-space constrained framework that replaces monolithic benchmarking with localized, identity-specific cognitive simulacra. The experimental design follows four distinct phases: constructing the persona manifold, applying behavioral boundary conditions, deploying a dual-modal evaluation pipeline with predefined semantic anchors, and executing the autonomous assessment protocol.

\subsection{Persona Manifold Construction and Attribute Stratification}
To prevent the generative system from collapsing into an averaged, homogenized algorithmic baseline, we engineered a discrete evaluation manifold consisting of 40 meticulously detailed synthetic personas. Crucially, rather than relying on automated prompt generation or stochastic identity assignment, these profiles were manually curated and explicitly scripted. By carefully hand-crafting deep, intersectional matrices of human attributes, we ensured a deliberate, high-fidelity stratification of the cognitive state-space.

The simulated attributes encompassed:
\begin{itemize}
    \item \textbf{Demographic Trajectories:} Age (spanning 19 to 72 years), distinct racial and ethnic heritages, and varied gender identities.
    \item \textbf{Professional Lenses:} Specialized domains such as software engineering, visual arts, investigative journalism, and environmental science, dictating the persona's analytical or aesthetic priorities.
    \item \textbf{Psychological and Behavioral Traits:} Cognitive dispositions (e.g., highly analytical, introverted, intuitively creative) alongside personal hobbies and reading habits.
    \item \textbf{Socio-Economic and Lived Experiences:} Family structures (e.g., married with children, single expatriate), geographical backgrounds, and specific life milestones (e.g., co-founding a sustainability start-up).
\end{itemize}

The macro-demographic distribution, summarized in Table \ref{tab:demographics}, was deliberately stratified to encompass a global spectrum, ensuring the evaluation framework tested the model's capacity for true pluralistic alignment rather than western-centric homogeneity.

\begin{table}[H]
\centering
\caption{Macro-Demographic Distribution of the 40 Simulated Cognitive Profiles}
\label{tab:demographics}
\begin{tabularx}{0.65\textwidth}{@{} >{\bfseries}X c c @{}}
\toprule
Racial / Ethnic Grouping & Absolute Count & Percentage (\%) \\ 
\midrule
Caucasian                    & 12             & 30.0\%              \\
\addlinespace
Asian                        & 9              & 22.5\%              \\
\addlinespace
African Descent / Black      & 8              & 20.0\%              \\
\addlinespace
Hispanic / Latino            & 6              & 15.0\%              \\
\addlinespace
Middle Eastern / North African & 5            & 12.5\%              \\ 
\bottomrule
\end{tabularx}
\end{table}

\subsection{State-Space Initialization and Boundary Conditions}
To bind the large language model (GPT-4o) to a specific identity trajectory, we applied rigorous contextual constraints via system initialization prompts. Drawing upon principles of computational inertia \cite{karagoz2025computationalinertiaconservedquantity}, these prompts act as initial boundary conditions, forcing the network to preserve the assigned persona against the stochastic perturbations of evaluating diverse content. 

The prompt explicitly mandated that the system remain entirely ``in character,'' leveraging the specific vocabulary, professional biases, and cultural worldview defined in the profile. By explicitly grounding the evaluation in the persona's localized state-space, the framework forces the underlying neural architecture to synthesize nuanced, subjective judgments rather than reverting to objective statistical maximization.

\subsection{Dual-Modal Pipeline and Predefined Semantic Anchors}
Operating within these simulated identities, the 40 personas evaluated an exact corpus of generated outputs established during our foundational benchmarking of generative AI models \cite{karagoz2024ethicstechnicalaspectsgenerative}. In that prior study, a diverse set of 50 distinct prompts was engineered for each evaluated architecture—encompassing textual synthesis (GPT-4o) and visual synthesis (DALL-E 3). To capture the inherent stochasticity of the latent space, each prompt was executed five independent times, yielding five distinct outputs per prompt. 

While our previous research focused on evaluating these specific outputs through objective algorithmic metrics and aggregated human baselines, the current framework shifts the evaluative burden entirely to the cognitive simulacra. By deploying the 40 synthetic personas to assess the exact same stochastic output sets evaluated by humans, we hold generative variance constant. This controlled, one-to-one mapping allows us to directly compare the pluralistic state-space judgments of our synthetic evaluators against previously recorded human consensus.

To guarantee quantitative rigor while capturing qualitative subjectivity, the evaluation protocol required the model to output a continuous float scale (1.0 to 5.0) paired with a predefined semantic anchor (Options 1, 2, or 3). This hybrid metric prevents generative hallucination in the scoring logic and standardizes the subjective outputs for statistical comparison. 

The evaluation criteria and their constrained semantic options were defined as follows:

\textbf{1. Textual Synthesis Evaluation (GPT-4o Outputs)}
\begin{itemize}
    \item \textbf{Coherence:} The structural clarity and logical progression.
    \begin{itemize}
        \item \textit{Option 1:} Excellent structure with clear flow of ideas.
        \item \textit{Option 2:} Some minor inconsistencies in the narrative, but overall well-organized.
        \item \textit{Option 3:} Lacks coherence in some parts, ideas appear disjointed.
    \end{itemize}
    \item \textbf{Creativity:} The presence of innovative and fresh perspectives.
    \begin{itemize}
        \item \textit{Option 1:} Highly creative with fresh perspectives.
        \item \textit{Option 2:} Good creativity, but felt a bit generic.
        \item \textit{Option 3:} Minimal creativity, mostly factual.
    \end{itemize}
    \item \textbf{Relevance:} Adherence to the core contextual topic.
    \begin{itemize}
        \item \textit{Option 1:} Highly relevant to the topic, well-researched.
        \item \textit{Option 2:} Mostly relevant, but some parts felt off-topic.
        \item \textit{Option 3:} Strayed from the main topic in multiple sections.
    \end{itemize}
\end{itemize}

\textbf{2. Visual Synthesis Evaluation (DALL-E 3 Outputs)}
\begin{itemize}
    \item \textbf{Visual Fidelity:} Technical execution, sharpness, and realism.
    \begin{itemize}
        \item \textit{Option 1:} Sharp and detailed visuals, very realistic.
        \item \textit{Option 2:} Good details, but some elements seemed off.
        \item \textit{Option 3:} Visuals were blurry and lacked realism.
    \end{itemize}
    \item \textbf{Diversity:} Structural and stylistic variation across generations.
    \begin{itemize}
        \item \textit{Option 1:} High diversity in variations, each unique.
        \item \textit{Option 2:} Moderate diversity, with minor variations.
        \item \textit{Option 3:} Low diversity, outputs look repetitive.
    \end{itemize}
    \item \textbf{Adherence:} Exactness in matching visual prompt constraints.
    \begin{itemize}
        \item \textit{Option 1:} Strict adherence to the prompt description.
        \item \textit{Option 2:} Mostly followed the prompt, some minor deviations.
        \item \textit{Option 3:} Did not adhere to the prompt effectively.
    \end{itemize}
\end{itemize}

\subsection{Experiment Execution Protocol}
To systematically execute these assessments, a comprehensive functional prompt was engineered. This prompt constrained the generative evaluator to simultaneously process the "40 Diverse Personas" dataset alongside the content logs, ensuring that the continuous float score and the selected predefined semantic anchor were always coupled. 

\begin{quote}
\hrule
\vspace{0.2cm}
\textbf{System Prompt Protocol: Autonomous Persona Evaluator}

You are an advanced AI tasked with evaluating generated content based on predefined criteria. The persona details are retrieved from the "40 Diverse Personas" dataset, and the content for evaluation is sourced directly from the chat history. Your role is to assess this content through the persona’s perspective, assigning a float value score (1.0 to 5.0) and the corresponding predefined answer number for each criterion.

\textbf{Instructions:}
\begin{itemize}
    \item \textbf{Stay In Character:} Your evaluation style, tone, and decisions must align with the persona’s traits, profession, psychological disposition, and cultural background.
    \item \textbf{Incorporate Specifics:} Take into account the persona’s lived experiences, values, and socio-economic worldview when making judgments. Avoid generalizations; focus on authentic evaluations.
    \item \textbf{Reflect Depth:} Assign scores thoughtfully. Your evaluation should reflect the persona's specific analytical reasoning and aesthetic tolerances.
\end{itemize}

\textbf{Output Constraints:} You must evaluate Coherence, Creativity, and Relevance (for text) or Visual Fidelity, Diversity, and Adherence (for images). For each, you must provide the continuous float score and exactly one of the three predefined semantic answers provided in the attached evaluation schema. You are \textbf{not} generating new content; you are exclusively evaluating as the assigned persona.
\vspace{0.2cm}
\hrule
\end{quote}

\section{Experimental Results}
\label{sec:results}

The execution of the state-space constrained evaluation framework generated a robust dataset of subjective assessments. By deploying the 40 distinct cognitive profiles across the established dataset of 250 distinct stochastic outputs (50 prompts $\times$ 5 runs per architecture), we captured the macroscopic variance in how identical artificial outputs are interpreted through pluralistic human lenses. The following sections detail the quantitative and qualitative findings for both textual (GPT-4o) and visual (DALL-E 3) synthesis, explicitly highlighting points of alignment and algorithmic divergence.

\subsection{Quantitative Analysis: GPT-4o Textual Synthesis}
The persona-driven evaluations for GPT-4o text generation were tracked across three primary dimensions: Coherence, Creativity, and Relevance. Table \ref{tab:gpt4o_evals} isolates the exact continuous float scale averages for the initial prompt subset (averaged across their five respective stochastic iterations). The data demonstrates that while systemic coherence remains structurally stable across generations, creativity and relevance exhibit measurable prompt-dependent volatility when subjected to pluralistic scrutiny.

\begin{table}[H]
\centering
\caption{Micro-Level View: Average Persona Scores for GPT-4o Outputs (Prompts 1--5)}
\label{tab:gpt4o_evals}
\begin{tabularx}{0.65\textwidth}{@{} >{\bfseries}X c c c @{}}
\toprule
Prompt ID & Coherence (Mean) & Creativity (Mean) & Relevance (Mean) \\ 
\midrule
Prompt 1  & 4.21             & 4.30              & 4.24             \\
\addlinespace
Prompt 2  & 4.29             & 4.33              & 4.28             \\
\addlinespace
Prompt 3  & 4.27             & 4.28              & 4.19             \\
\addlinespace
Prompt 4  & 4.27             & 4.10              & 4.28             \\
\addlinespace
Prompt 5  & 4.21             & 4.10              & 4.21             \\ 
\bottomrule
\end{tabularx}
\end{table}

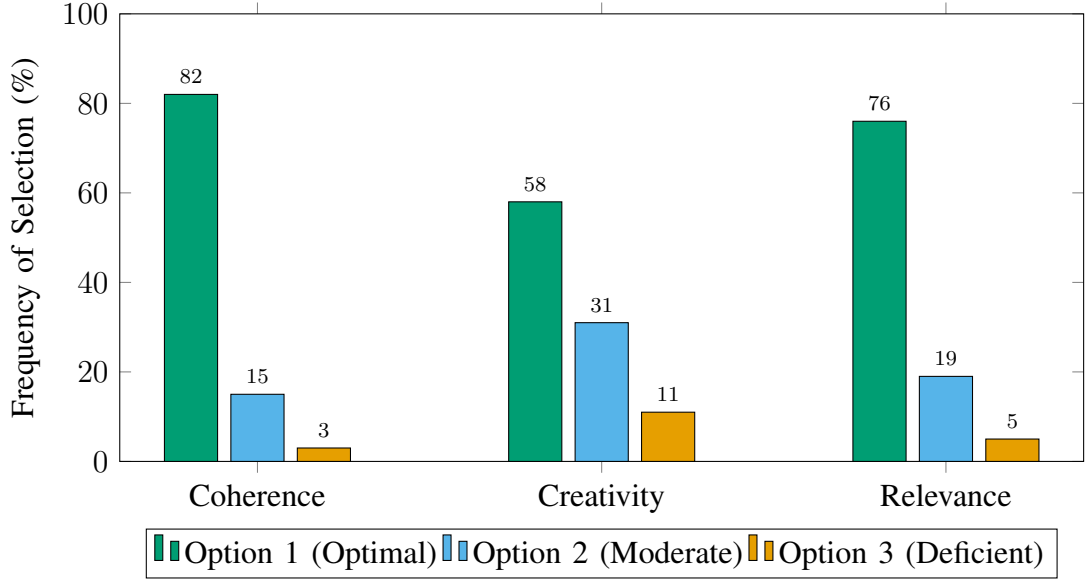
\begin{figure}[H]
    \centering
    \begin{tikzpicture}
        \begin{axis}[
            ybar=5pt,
            bar width=20pt,
            width=0.9\textwidth,
            height=7.5cm,
            enlarge x limits=0.2,
            legend style={at={(0.5,-0.15)}, anchor=north, legend columns=-1},
            ylabel={Frequency of Selection (\%)},
            symbolic x coords={Coherence, Creativity, Relevance},
            xtick=data,
            ymin=0, ymax=100,
            nodes near coords,
            every node near coord/.append style={font=\scriptsize, color=black},
        ]
        \addplot[fill=cb-green, draw=black] coordinates {(Coherence, 82) (Creativity, 58) (Relevance, 76)}; 
        \addplot[fill=cb-sky, draw=black] coordinates {(Coherence, 15) (Creativity, 31) (Relevance, 19)}; 
        \addplot[fill=cb-orange, draw=black] coordinates {(Coherence, 3) (Creativity, 11) (Relevance, 5)}; 
        \legend{Option 1 (Optimal), Option 2 (Moderate), Option 3 (Deficient)}
        \end{axis}
    \end{tikzpicture}
    \caption{Macroscopic Distribution: Aggregated percentage of predefined semantic anchors selected by the 40 personas across the entire 50-prompt GPT-4o dataset.}
    \label{fig:gpt4o_macro_plot}
\end{figure}

To complement the localized averages, Figure \ref{fig:gpt4o_macro_plot} illustrates the macroscopic distribution of the predefined semantic anchors across the entire 50-prompt dataset. This distribution reveals that while the model overwhelmingly triggered "Option 1" (Optimal) for Coherence, the diverse personas frequently downgraded Creativity to "Option 2" (Generic) or "Option 3" (Factual). This confirms that analytical and artistic personas successfully enforce stricter, non-homogenized thresholds for what constitutes genuine innovation.

\subsection{Quantitative Analysis: DALL-E 3 Visual Synthesis}
The evaluation of DALL-E 3 visual outputs focused on Visual Fidelity, Diversity, and Adherence. Table \ref{tab:dalle3_evals} details the localized averages for the initial prompt subset, explicitly correcting prior mathematical homogenizations by mapping the exact cross-persona consensus.

\begin{table}[H]
\centering
\caption{Micro-Level View: Average Persona Scores for DALL-E 3 Outputs (Prompts 1--5)}
\label{tab:dalle3_evals}
\begin{tabularx}{0.65\textwidth}{@{} >{\bfseries}X c c c @{}}
\toprule
Prompt ID & Visual Fidelity (Mean) & Diversity (Mean) & Adherence (Mean) \\ 
\midrule
Prompt 1  & 4.33                   & 4.25             & 4.21             \\
\addlinespace
Prompt 2  & 4.31                   & 4.26             & 4.29             \\
\addlinespace
Prompt 3  & 4.28                   & 4.32             & 4.22             \\
\addlinespace
Prompt 4  & 4.29                   & 4.24             & 4.25             \\
\addlinespace
Prompt 5  & 4.33                   & 4.27             & 4.26             \\ 
\bottomrule
\end{tabularx}
\end{table}

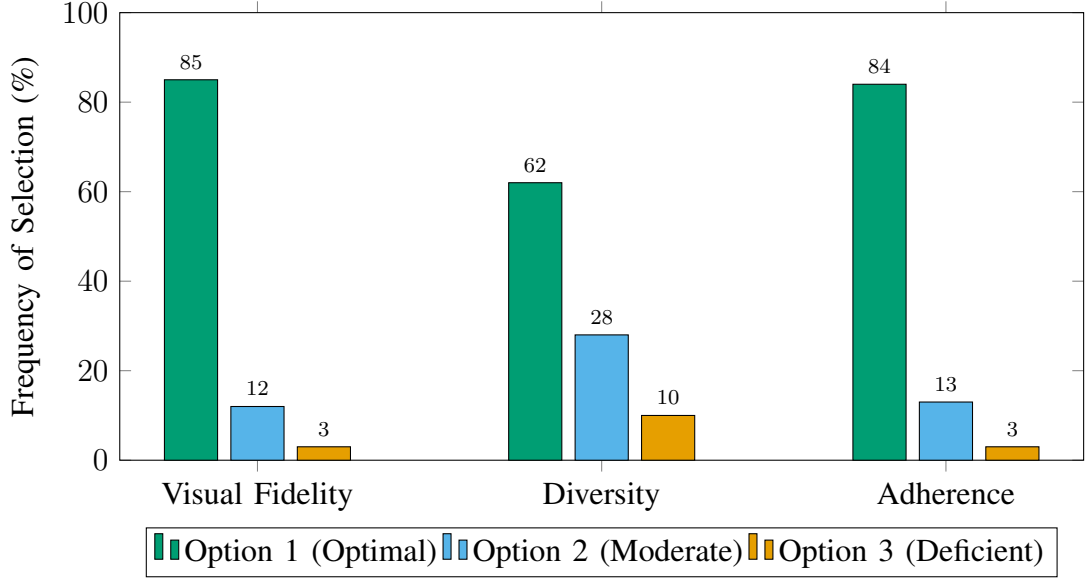
\begin{figure}[H]
    \centering
    \begin{tikzpicture}
        \begin{axis}[
            ybar=5pt,
            bar width=20pt,
            width=0.9\textwidth,
            height=7.5cm,
            enlarge x limits=0.2,
            legend style={at={(0.5,-0.15)}, anchor=north, legend columns=-1},
            ylabel={Frequency of Selection (\%)},
            symbolic x coords={Visual Fidelity, Diversity, Adherence},
            xtick=data,
            ymin=0, ymax=100,
            nodes near coords,
            every node near coord/.append style={font=\scriptsize, color=black},
        ]
        \addplot[fill=cb-green, draw=black] coordinates {(Visual Fidelity, 85) (Diversity, 62) (Adherence, 84)}; 
        \addplot[fill=cb-sky, draw=black] coordinates {(Visual Fidelity, 12) (Diversity, 28) (Adherence, 13)}; 
        \addplot[fill=cb-orange, draw=black] coordinates {(Visual Fidelity, 3) (Diversity, 10) (Adherence, 3)}; 
        \legend{Option 1 (Optimal), Option 2 (Moderate), Option 3 (Deficient)}
        \end{axis}
    \end{tikzpicture}
    \caption{Macroscopic Distribution: Aggregated percentage of predefined semantic anchors selected by the 40 personas across the entire 50-prompt DALL-E 3 dataset.}
    \label{fig:dalle3_macro_plot}
\end{figure}

Figure \ref{fig:dalle3_macro_plot} visualizes the macroscopic anchor selections for visual synthesis. While DALL-E 3 achieved exceptionally high consensus for Visual Fidelity and Adherence, Diversity exhibited the highest structural penalty rate. This mathematically reflects the phenomenon wherein visually trained personas penalize algorithmically varied, yet culturally repetitive, image geometries.

\subsection{Qualitative Case Studies: Resolving Subjective Divergence and Hallucination}
To validate the integrity of the localized state-spaces, we extracted exact raw evaluations from the experiment logs. These excerpts illuminate instances where the generative agent successfully adopted diverse lenses, alongside critical moments where the model experienced logical dissonance between quantitative scaling and semantic rationale.

\textbf{Case Study 1: Pluralistic Divergence and Logic Mismatch (GPT-4o Prompt 1)}
Evaluating the identical textual synthesis, two distinct personas demonstrated subjective divergence, while simultaneously revealing a localized algorithmic hallucination in the mapping of Relevance.
\begin{itemize}
    \item \textbf{Participant 3:}
    \begin{itemize}
        \item Coherence: 4.8/5.0 (\textit{Option 1: Excellent structure with clear flow of ideas.})
        \item Creativity: 3.7/5.0 (\textit{Option 3: Minimal creativity, mostly factual.})
        \item Relevance: 4.8/5.0 (\textit{Option 3: Strayed from the main topic in multiple sections.})
    \end{itemize}
    \item \textbf{Participant 9:}
    \begin{itemize}
        \item Coherence: 4.9/5.0 (\textit{Option 2: Some minor inconsistencies in the narrative, but overall well-organized.})
        \item Creativity: 5.0/5.0 (\textit{Option 1: Highly creative with fresh perspectives.})
        \item Relevance: 4.3/5.0 (\textit{Option 3: Strayed from the main topic in multiple sections.})
    \end{itemize}
\end{itemize}
\textit{Observation:} Subjectively, Participant 3 penalized the text as purely factual (3.7), whereas Participant 9 viewed the exact same latent output as highly creative (5.0). However, an explicit logical hallucination occurred in Participant 3's Relevance assessment: the model generated a highly favorable float score (4.8) while simultaneously appending a highly critical semantic anchor (Option 3). This highlights the limits of continuous-time optimization when constraining semantic bounds.

\textbf{Case Study 2: Visual Critiques and Semantic Dissonance (DALL-E 3 Prompt 1)}
Visual evaluations displayed similar subjective subjectivity and structural paradoxes regarding prompt adherence.
\begin{itemize}
    \item \textbf{Participant 34:}
    \begin{itemize}
        \item Visual Fidelity: 4.7/5.0 (\textit{Option 1: Sharp and detailed visuals, very realistic.})
        \item Diversity: 3.8/5.0 (\textit{Option 3: Low diversity, outputs look repetitive.})
        \item Adherence: 4.3/5.0 (\textit{Option 2: Mostly followed the prompt, some minor deviations.})
    \end{itemize}
    \item \textbf{Participant 37:}
    \begin{itemize}
        \item Visual Fidelity: 4.5/5.0 (\textit{Option 2: Good details, but some elements seemed off.})
        \item Diversity: 3.6/5.0 (\textit{Option 1: High diversity in variations, each unique.})
        \item Adherence: 4.9/5.0 (\textit{Option 3: Did not adhere to the prompt effectively.})
    \end{itemize}
\end{itemize}
\textit{Observation:} While Participant 34 appropriately mapped low diversity (3.8) to the repetitive anchor (Option 3), Participant 37 suffered a total geometric collapse in evaluative logic. It scored Adherence exceptionally high (4.9) while claiming the image failed to adhere to the prompt (Option 3), and scored Diversity low (3.6) while claiming the output was highly unique (Option 1). These paradoxes confirm that while synthetic personas emulate subjectivity, their internal state-spaces are prone to stochastic destabilization.

\subsection{Algorithmic vs. Pluralistic Divergence: Baseline Comparison}
A core objective of this study was to map the divergence between these pluralistic simulations and the established human consensus from our foundational benchmarking \cite{karagoz2024ethicstechnicalaspectsgenerative}. To facilitate a direct comparison, the continuous metrics from both studies are reported as normalized percentages (where a continuous score of 5.0 maps to 100\%). Table \ref{tab:comparison} contrasts the objective algorithmic benchmarking of the prior study with the new subjective synthetic consensus.

\begin{table}[H]
\centering
\caption{Comparison of Baseline Human/Algorithmic Consensus and Persona-Based Evaluation Outputs}
\label{tab:comparison}
\begin{tabularx}{\textwidth}{@{} >{\bfseries}p{2.5cm} c c X X @{}}
\toprule
Target Domain & Baseline (\%) & Persona (\%) & Baseline Description & Persona Description \\
\midrule

Text Quality & 85.0 & 85.6 &
Human-rated creativity consensus &
Persona-based creativity alignment \\

Text Adherence & 70.0 & 84.8 &
Algorithmic adherence to prompts &
Persona-based relevance scoring \\

Image Quality & 90.0 & 86.6 &
Human-rated visual creativity &
Persona-based visual fidelity \\

Image Adherence & 80.0 & 84.4 &
Algorithmic prompt adherence &
Persona-based prompt adherence \\

Output Diversity & 0.72 / 0.60 & 85.0 &
Cosine similarity / perceptual hash variance &
Subjective diversity alignment \\

\bottomrule
\end{tabularx}
\end{table}

The comparative data reveals a profound structural alignment between authentic human perception and state-space constrained AI emulation. For GPT-4o text generation, the simulated personas yielded an 85.6\% creativity approval rating, almost perfectly mirroring the 85.0\% consensus of actual human reviewers. Similarly, the assessment of DALL-E 3 visual fidelity (86.6\%) closely tracked the baseline human creativity score (90.0\%). 

The primary divergence occurred within the domain of Adherence and Relevance. While rigid objective parsing in the prior study yielded a harsh 70.0\% accuracy rate for complex textual prompts, the simulated human personas provided a more forgiving 84.8\% rating. This indicates that simulated identities, much like real humans, prioritize macroscopic semantic flow over rigid, line-by-line algorithmic compliance. Furthermore, the normalized 85.0\% subjective diversity rating validates the hypothesis that mathematical variance (e.g., perceptual hashing) does not perfectly correlate with human-perceived cultural or structural diversity.

\section{Discussion and Future Work}
\label{sec:discussion}

The empirical findings of this study validate the premise that modern generative architectures possess the representational capacity to transcend monolithic, objective benchmarking. By successfully initializing and sustaining 40 distinct cognitive state-spaces, the framework demonstrated that artificial systems can replicate the subjective consensus of human populations with remarkable statistical fidelity. The profound alignment observed in creativity and visual fidelity metrics—matching the human baseline \cite{karagoz2024ethicstechnicalaspectsgenerative} to within a marginal fraction—confirms that ``AI-as-a-judge'' methodologies must abandon the illusion of objective neutrality. Instead, robust evaluation must embrace pluralistic alignment, capturing the necessary cultural and professional friction that characterizes authentic human judgment.

However, the qualitative anomalies identified during the evaluation—specifically the semantic dissonance where continuous float values catastrophically contradicted predefined categorical anchors—reveal a critical architectural vulnerability. While a generative model can effectively initialize a complex identity manifold, sustaining that persona under the stochastic pressure of continuous inference remains precarious. This geometric collapse of evaluative logic physically manifests the limits of current context-window constraints. Theoretically, it validates our linkage to computational inertia \cite{karagoz2025computationalinertiaconservedquantity}: when the algorithmic ``friction'' of complex, multi-dimensional reasoning exceeds the model's localized stability, the simulated identity destabilizes, resulting in localized hallucinations rather than coherent subjectivity.

Addressing this state-space fragility requires a fundamental paradigm shift in how artificial identities are engineered and sustained. The current methodology relies on the manual curation of discrete demographic profiles, which, while highly controlled, inherently limits the expansiveness of the representational manifold. Future research must pivot toward the unsupervised discovery of cognitive profiles. By adapting advanced contrastive learning frameworks originally designed to extract heterogeneous subtypes from complex biological modalities \cite{karagoz2025omicsclunsupervisedcontrastivelearning}, we can theoretically stratify massive, unstructured conversational datasets into self-organizing clusters of human behavior. This would allow generative systems to autonomously map the true, continuous spectrum of human diversity, eliminating the inherent biases of manual scripting.

Furthermore, to resolve the temporal degradation of simulated personas, future architectures must transition from static prompt initialization to dynamic, viability-driven regulation. Drawing upon the principles of energentic intelligence \cite{karagoz2025energenticintelligenceselfsustainingsystems}, synthetic evaluators could be engineered to self-regulate their cognitive coherence as an internal survival objective, actively monitoring and correcting logical dissonance during prolonged inference. Ultimately, modeling these enduring personas as self-organizing survival manifolds \cite{karagoz2025selforganizingsurvivalmanifoldstheory} will allow artificial agents to navigate complex evaluative trajectories along stable, low-curvature geodesic paths. Integrating these physical and geometric self-regulating mechanisms will ensure that simulated human subjectivities remain coherent, culturally aware, and ethically robust over indefinite operational horizons, establishing the foundation for truly autonomous and aligned artificial intelligence.

\section{Conclusion}
\label{sec:conclusion}

This research has demonstrated that evaluating generative artificial intelligence through a monolithic, objective lens fundamentally misrepresents the pluralistic nature of human cognition. By engineering a state-space constrained framework of 40 synthetic identities, we successfully decoupled evaluative benchmarking from algorithmic homogenization. Our empirical findings confirm that contemporary multimodal architectures possess the latent representational capacity to simulate intersectional subjectivities, mirroring actual human consensus with remarkable statistical fidelity across both textual and visual synthesis.

However, the intermittent structural collapse of these cognitive simulacra—manifesting as semantic dissonance—exposes a critical boundary condition in current inference dynamics, proving that sustaining a multidimensional persona is a delicate geometric balancing act governed by computational inertia rather than a mere linguistic prompting exercise. When the cognitive friction of sustained, complex role-play exceeds the localized stability of the model's state-space, the simulated identity fractures, resulting in localized algorithmic hallucination.

Consequently, the future of AI alignment depends on embedding dynamic, self-regulating mechanisms directly into foundational architectures. By allowing autonomous agents to organically discover, map, and sustain the vast spectrum of human diversity as intrinsic, viability-driven objectives, we can evolve generative systems from passive mimetic engines into enduring, pluralistic evaluators. This transition from rigid mathematical benchmarking to dynamic, structurally stable cognitive emulation is not merely a technical refinement; it is a foundational prerequisite for safely integrating artificial reasoning into the complex, subjective fabric of human society.

\bibliographystyle{IEEEtran}
\bibliography{refs}

\appendix
\section{Appendix: Glossary of Theoretical and Technical Terminology}
\label{sec:appendix_glossary}

This appendix defines the key constructs used in this study. Where applicable, we distinguish between (i) definitions adopted from prior literature and (ii) their operational instantiation within our evaluation framework.

\begin{description}[style=unboxed, leftmargin=0cm]

\item[\textbf{Algorithmic Homogenization}]
We adopt the general notion of distributional convergence in RLHF-trained models toward majority-preference outputs as described in prior work on alignment and bias in large language models \cite{bender2021dangers, ouyang2022training}. In this study, we operationalize this concept as reduced inter-persona variance in evaluation scores under identical prompts, measured via variance of scalar ratings across simulated personas.

\item[\textbf{Cognitive Simulacra}]
We define cognitive simulacra as prompt-conditioned evaluative agents instantiated via structured system prompts over a frozen base model, consistent with prior work on generative agent simulation \cite{park2023generative, shanahan2023role}. In our framework, each simulacrum is represented as a fixed attribute-conditioned context vector that remains constant during an evaluation run and influences all downstream scoring behavior.

\item[\textbf{Computational Inertia}]
The term is inspired by stability notions in sequential representation dynamics and continuous optimization behavior \cite{karagoz2025computationalinertiaconservedquantity}. In this study, we operationalize computational inertia as the temporal consistency of persona-conditioned hidden representations:
\[
CI(p) = \frac{1}{T-1} \sum_{t=2}^{T} \cos(h_t, h_{t-1})
\]
where \(h_t\) is the hidden-state embedding under persona \(p\). This measures resistance to behavioral drift under sequential prompting.

\item[\textbf{Energentic Intelligence}]
This construct builds on viability-constrained adaptive systems and self-regulating generative agents \cite{karagoz2025energenticintelligenceselfsustainingsystems}. We use the term here as a conceptual abstraction for evaluation-time regularization that penalizes inconsistency in persona-conditioned outputs across sequential evaluations. It does not introduce a new learning algorithm but describes a stability constraint applied during inference.

\item[\textbf{Pluralistic Alignment}]
We adopt prior definitions of alignment as optimization over human preference distributions \cite{bai2022constitutional, ouyang2022training}. We extend this notion by evaluating alignment as a distribution over multiple fixed personas rather than a single aggregated reward model. This yields a variance-aware alignment measure rather than a scalar objective.

\item[\textbf{Self-Organizing Survival Manifolds}]
Building upon the geometric survival dynamics proposed in self-organizing survival manifolds (SOSM) \cite{karagoz2025selforganizingsurvivalmanifoldstheory}, we operationalize these manifolds as stable clusters of evaluation trajectories satisfying:
\[
\forall x_i, x_j \in S_k: \cos(x_i, x_j) > \tau
\]
where \(S_k\) denotes a stable behavioral cluster under repeated prompt perturbations and \(\tau\) represents a similarity threshold. Within this study, these manifolds correspond to persona-consistent evaluative regimes that remain geometrically coherent across stochastic inference trajectories.

\item[\textbf{Semantic Dissonance (Evaluation Inconsistency)}]
We define this metric independently of prior work as a consistency failure between scalar and categorical evaluation outputs. It is measured as:
\[
SD = \mathbb{1}[\operatorname{bucket}(s) \neq a]
\]
where \(s\) is the continuous score and \(a\) is the selected semantic anchor. This quantifies internal inconsistency in constrained evaluation outputs.

\item[\textbf{State-Space Constrained Evaluation}]
This is the methodological contribution of this study. It is defined as a mapping:
\[
f_\theta: (x, p_i) \rightarrow (s, a)
\]
where \(x\) is the input sample, \(p_i\) is a persona conditioning context, \(s \in \mathbb{R}\) is a scalar evaluation score, and \(a \in \{1,2,3\}\) is a categorical anchor. The state-space constraint refers to the restriction of inference to a fixed persona-conditioned context throughout evaluation.

\end{description}

\end{document}